\documentclass{llncs}

\usepackage{makeidx}
\usepackage{multirow}
\usepackage{amsmath}
\usepackage{graphicx}

\begin{document}

\title{General Purpose Textual Sentiment Analysis and Emotion Detection Tools}
\titlerunning{Sentiment Analysis}
\author{Alexandre Denis \and Samuel Cruz-Lara \and Nadia Bellalem}
\authorrunning{Alexandre Denis et al.}
\tocauthor{Alexandre Denis, Samuel Cruz-Lara, Nadia Bellalem}
\institute{LORIA UMR 7503, SYNALP Team\\ University of Lorraine, Nancy, France \\
\email{\{alexandre.denis, samuel.cruz-lara, nadia.bellalem\}@loria.fr}}

\maketitle 

\begin{abstract}

Textual sentiment analysis and emotion detection consists in retrieving the sentiment or emotion carried by a text or
document. This task can be useful in many domains: opinion mining, prediction, feedbacks, etc. However, building a
\emph{general purpose} tool for doing sentiment analysis and emotion detection raises a number of issues, theoretical
issues like the dependence to the domain or to the language but also pratical issues like the emotion representation
for interoperability. In this paper we present our sentiment/emotion analysis tools, the way we propose to circumvent
the difficulties and the applications they are used for.

\end{abstract}

%%%%%%%%%%%%%%%%%%%%%%%%%%%%%%%%%%%%%%%%%%%%%%%%%%%%%%%%%%%%%%%%
\section{Sentiment Analysis and Emotion Detection from text}

\subsection{Definition}

One of the most complete definition of the sentiment analysis task is proposed by Liu \cite{liu2012} in which a
sentiment is defined as a quintuple $\langle e, a, s, h, t\rangle$ where $e$ is the name of an entity, $a$ an
aspect of this entity, $s$ is a sentiment value about $a$, $h$ is the opinion holder and $t$ is the time when the
opinion is expressed by $h$. The sentiment analysis task consists in outputing for a given text or sentence the set
of sentiments that this text conveys. Whereas sentiment analysis is limited in general to binary sentiment value
(positive, negative) or ternary (positive, negative, neutral), emotion detection consists in determining the
sentiment value among a larger set of emotions, typically Ekman's emotions \cite{ekman1972}: joy, fear, sadness,
anger, disgust, or surprise.

\subsection{Difficulties and existing approaches}

A difficult aspect of sentiment analysis is the fact that a given word can have a different polarity in a different
context. In \cite{wilson2009}, authors oppose the \emph{prior polarity} of a word, that is the polarity that a word
can have out of context, and the \emph{contextual polarity}, that is the polarity of a word in a particular
context. There are many complex phenomena that influence the contextual valence of a word
\cite{polanyi2004,wilson2009,liu2012} as the table~\ref{table:phenomena} shows, and most of the approaches thus
reduce the scope of the problem to $\langle s \rangle$ the sole sentiment value or sometimes to $\langle e, s, h
\rangle$, the entity, the sentiment value and the holder like in \cite{kim2006}. 

\begin{figure}
\begin{small}
\begin{center}
\begin{tabular}{|l|l|l|}
\hline
\emph{Phenomenon} 			& \emph{Example} & \emph{Polarity} \\ \hline
\multirow{1}{*}{Negation}		& it's not good ; no one thinks it is good & negative \\ \hline
\multirow{1}{*}{Irrealis}	& it would be good if... ; if it is good then ... & neutral \\ \hline
\multirow{2}{*}{Presupposition}	& how to fix this terrible printer? & negative \\ 
						& can you give me a good advice?& neutral\\ \hline
\multirow{1}{*}{Word sense}			& this camera sucks ; this vaccum cleaner sucks & negative vs positive \\ \hline
Point of view			& Israel failed to defeat Hezbollah & negative or positive \\ \hline
Common sense			& this washer uses a lot of water & negative \\ \hline
\multirow{2}{*}{Multiple entities}		& \multirow{2}{*}{Ann hates cheese but loves cheesecake} & negative wrt cheese \\ 
						&					 & positive wrt cheesecake \\ \hline
\multirow{2}{*}{Multiple aspects}		& \multirow{2}{*}{this camera is awesome but too expensive} & positive wrt camera \\
						& 					   & negative wrt price \\ \hline
\multirow{2}{*}{Multiple holders}		& \multirow{2}{*}{Ann hates cheese but Bob loves it} & negative wrt Ann \\
						&				    & positive wrt Bob \\
\hline
\multirow{2}{*}{Multiple time}			& \multirow{2}{*}{Ann used to hate cheese and now she loves it} & negative wrt past \\
						&  					       & positive wrt present \\ \hline
\end{tabular}
\end{center}
\end{small}
\caption{Examples of linguistic phenomena that influence the final valence of a text}
\label{table:phenomena}
\end{figure}

Given the complexity of the task, it is not a surprise that the first approaches to the problem use machine learning.
In \cite{turney2002} an unsupervised approach is proposed. It uses the Pointwise Mutual Information distance between
online reviews and the words ``excellent'' and ``poor''. Its accuracy ranges from 84\% for the automobile reviews to
66\% for the movie reviews. In \cite{pang2002}, a corpus of movie reviews and their annotation in terms of number of
stars is used to train several classifiers (Naive Bayes, Support Vector Machine and Maximum Entropy) and obtain a
good score for the best (around 83\%). More recent work in machine learning approaches to sentiment analysis explore
successfully different kinds of learning algorithms such as Conditional Random Fields \cite{nakagawa2010} or
autoencoders which are a kind of Neural Networks and which offer good performance on the movie reviews domain
\cite{socher2011}. A detailed and exhaustive survey of the field can be found in \cite{liu2012}.

%%%%%%%%%%%%%%%%%%%%%%%%%%%%%%%%%%%%%%%%%%%%%%%%%%%%%%%%%%%%%%%%
\section{Cross Domains problems}

A general purpose sentiment analysis or emotion detection tool is meant to work in different domains with different
applications and thus faces at least three problems: the dependence of the algorithms to the domain they were
developed on, the representation of emotions/sentiments for interoperability, and the fact that different
applications may require other languages than English, and as such the multilinguality issue must be considered. We
detail these three problems in this section.

\subsection{Domain dependence}

\paragraph{Machine learning dependence} An important issue of supervised machine learning is the dependence to the
training domain. Classical supervised algorithms require a new training corpus each time a new domain is tackled.
In \cite{aue2005} several methods are tried to overcome the domain-dependence of machine learning and they show
that the best results can be obtained by combining small amounts of labeled data from the training domain and large
amounts of unlabeled datas in the target domain. Actually, unsupervised or semi-supervised machine learning seems
more adequate than purely supervised machine learning to reach domain-independence \cite{read2009}. Another
approach \cite{andreevskaia2008} is to use hybrid methods, classifiers trained on corpora and polarity lexicons.
Indeed, polarity lexicons, such as Sentiwordnet \cite{esuli2006} or the Liu Lexicon \cite{hu2004}, being in general
domain-independent seem to be an interesting track to follow.

\paragraph{Types of emotions dependence} Moreover the set of relevant emotions depends on the domain. In a generic
emotion analysis tool, there is not much choice apart providing a set of the least domain specific emotions, hence
the frequent choice of Ekman's emotions. These may not describe accurately the affective states in all domains, for
instance their use is criticized in the learning domain~\cite{kort2001,baker2010}, but their independence to the
domain and the existence of Ekman's based emotional lexicons such as WordNet-Affect~\cite{strapparava2004} makes
them a common practical choice.

\subsection{Interoperability}

The representation of emotions for interoperability is an important issue for a sentiment/emotion analysis tool
that is meant to work in several domains and with several applications. We advocate the use of EmotionML
\cite{schroder2012} a W3C proposed recommendation for the representation of emotions. An interesting aspect of
EmotionML is the acknowledgement that there exists no consensus on how to represent an emotion, for instance is an
emotion better represented as a cognitive-affective state like \cite{kort2001}, as a combination between pleasure
and arousal like \cite{baker2010}, or as an Ekman emotion \cite{ekman1972}? Thus, EmotionML proposes instead an
emotion skeleton whose features are defined by the target application. An emotion is defined by a set of
descriptors, either dimensional (a value between 0 and 1) or categorical (a discreet value), and each descriptor
refers to an emotional vocabulary document. An emotion and its vocabulary can be embedded in one single document or
the emotion can refer to an online vocabulary document.

\subsection{Multilinguality}

A general purpose sentiment/emotion analysis tool is also required to be working in other languages than English.
Most of the work related to multilinguality is tied to subjectivity analysis, a simpler sentiment-like analysis
which consists in determining whether a text conveys an objective or subjective assessement. Several solutions are
possible, for instance training classifiers on translated corpora, using translated lexicons, building lexicons or
corpora for targeted language~\cite{banea2011}. Recent experiments with automatic translation for sentiment
analysis show that the performance of machine translation does not degrade the results too much \cite{balahur2012}.
We refer the interested reader to \cite{banea2011,liu2012} for a good overview of the topic.

%%%%%%%%%%%%%%%%%%%%%%%%%%%%%%%%%%%%%%%%%%%%%%%%%%%%%%%%%%%%%%%%
\section{Tools and applications}

We present here the sentiment/emotion analysis tools and the applications that use them in the context of the
Empathic Products ITEA2 project (11005)\footnote{http://www.empathic.eu/}, a european project dedicated to the
creation of applications that adapt to the intentional and emotional state of the users. 

\subsection{Sentiment/emotion analysis tools}

We implemented several sentiment analysis and emotion detection engines, and will briefly present them here. All of
them are integrated in one Web API that takes text in input and returns a single emotion formatted in EmotionML
format, dimensional valence for sentiment analysis engines and categorical emotion for the emotion detection
engines\footnote{The Web API is currently accessible on http://talc2.loria.fr/empathic}. A support also exists for
non-English languages following \cite{balahur2012} using Google machine translation but this service has not
yet been evaluated.

For emotion detection, the approach uses an emotion lexicon, namely WordNet-Affect \cite{strapparava2004} by
detecting emotional keywords in text complemented with a naive treatment of negation which inverts the found
emotion, ad hoc filters (smileys, keyphrases) and simple semantic rules. Despite its simplicity this tool manages
to reach performance similar to other approaches when evaluated on the Semeval-07 affective task dataset
\cite{strapparava2007}, it obtains 54.9\% of accuracy given an emotion to valence mapping (joy is positive and the
others are negative).

For sentiment analysis, we are currently exploring two opposed approaches, one symbolic and one with machine
learning. The symbolic approach follows \cite{polanyi2004} as an attempt to both tackle the linguistic difficulties
we mentioned thanks to valence shifting rules and domain dependence by using general purpose lexicons. It works by
first retrieving the prior polarity of words as found in the Liu lexicon \cite{hu2004} after a part-of-speech
tagging phase with the Stanford CoreNLP library. Then, a parsing phase enables to construct the dependencies (also
with Stanford CoreNLP), the resulting dependencies are filtered such that prior word valence is propagated along the
dependencies following manually crafted rules for valence shifting or inversion. This approach enables to be more
precise, for instance the sentence ``I don't think it's a missed opportunity'' would be tagged as positive by the
application of two valence flipping rules, a modifier rule for ``missed opportunity'', and a verb negation rule for
``don't think''. This approach obtains 56.3\% accuracy on the Semeval-07 dataset and 65.86\% accuracy on the
Semeval-13 data set. The impact of rules has also been evaluated and show that the rules enable to gain 5\% accuracy
on the Semeval-13 dataset as opposed to a simple lexical approach that only takes the average valence of all words
contained in a sentence.

The second approach relies on machine learning by training a classifier, namely a Random Forest classifier evaluated
on the Semeval-13 dataset. It uses simple features such as the stemmed words and the part-of-speech but manages to
obtain 64.30\% accuracy on Semeval-07 and 60.72\% accuracy on Semeval-13 both evaluated with 10-fold cross
validation. We also performed preliminary evaluation of the cross-domain abilities of the statistical approach and
observed that when trained on the Semeval-07 dataset and evaluated on Semeval-13 it obtains 47.08\% accuracy. While
training it on Semeval-13 and evaluating it on Semeval-07, it obtains 55.5\% accuracy. The significant difference is
likely caused by the dataset dissimilarities: first the Semeval-07 dataset is much smaller than Semeval-13 (1000
utterances vs 7500 utterances) and then, training on Semeval-07 is less efficient, and second the Semeval-07 dataset
only consists in short news headline while the Semeval-13 dataset is composed of tweets which are much longer and
then a better source for training. We assume that the difference in text length could also explain the difference in
the results for the symbolic approach (56\% vs 65.86\%) since it is known that text length can influence the
performance of sentiment analysis engines. A less difficult cross-domain evaluation would then rely on datasets that
share the same properties in terms of text length and available data size.

\subsection{Applications}

\subsubsection{Video conference feedback}

\begin{figure}
\centering
\includegraphics[width=2cm]{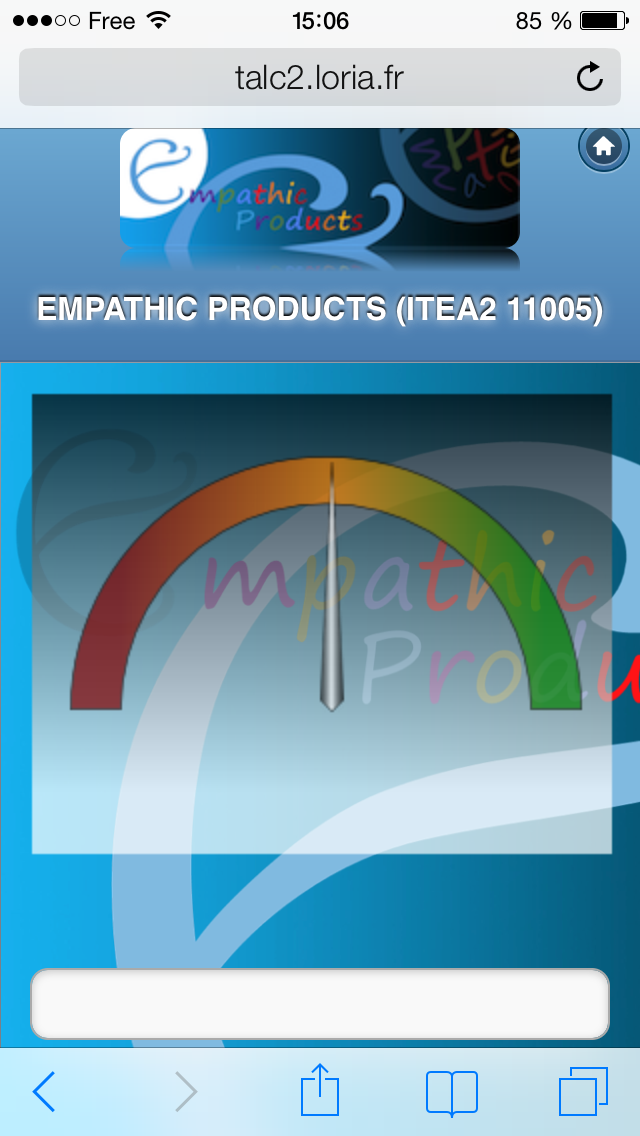}
\includegraphics[width=2cm]{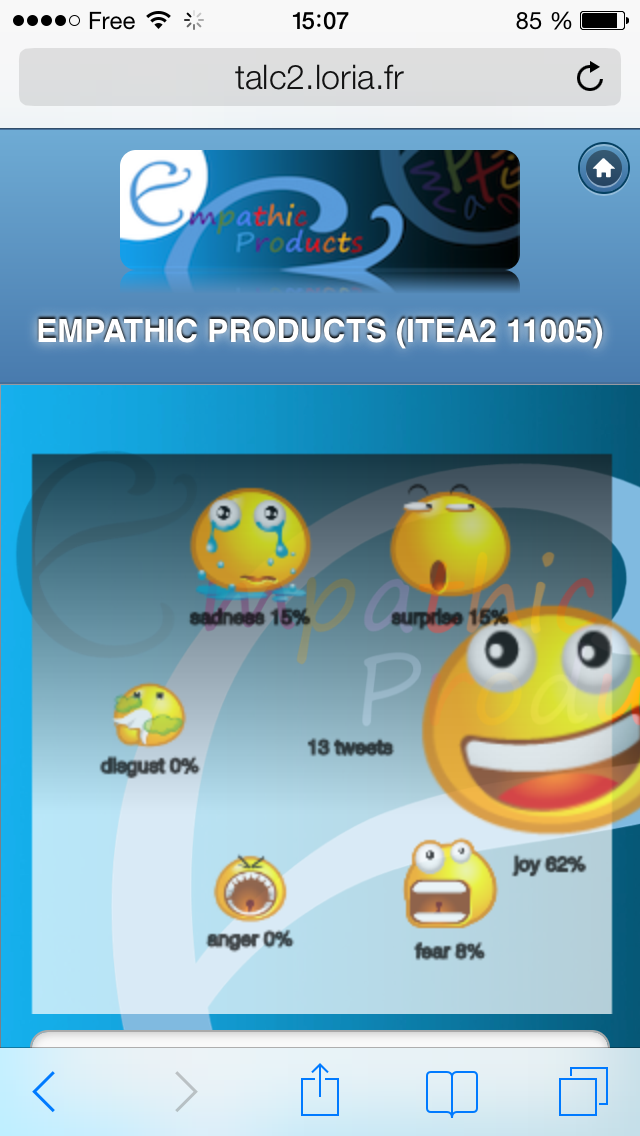}
\caption{Display of valence and emotions on mobile device}
\label{fig:gauge_mobile}
\end{figure}

One problem of video conference is the lack of feedback that the presenter can have about its remote audience. The
first application of our sentiment/emotion analysis tool consists in providing the presenter an aggregated feedback
of the emotional state of its audience. We assume that the audience is both attending to the videoconference
remotely and expressing its feelings over a textual channel, using Twitter, Facebook or by chat. Moreover in the
context of the Empathic Products project, the audience video feedback is also analyzed with regards to visual
emotions. The interoperability solution based on EmotionML proves to be an efficient option for combining the two
kinds of feedbacks. The most generic emotional output is the binary valence which offers a basic yet more reliable
characterization of the affective state of the audience. Emotions are also possible, but depending on the video
domain (e-learning, news, etc.), Ekman's emotions may not be fully relevant. The sentiment/emotion of all messages
sent by audience participants is averaged; when using valence it is possible to animate a gauge, when using
emotions, it is possible to animate iconic emotion representations (fig.~\ref{fig:gauge_mobile}) 

\subsubsection{E-learning virtual world emotion tagging}

E-learning in a virtual world requires some level of investment by the participants and is eased by their
collaboration. It has been shown that emotional information can enhance the collaboration. For instance
\cite{kamada2005} show that participants interact more if provided with emotional clues about their partner's current
state. We propose to integrate our sentiment analysis tool to the Umniverse collaborative virtual world
\cite{reis2011}: the virtual world works as a situated forum in which participants can move around and submit posts.
When participants submit posts they can annotate by hand the emotion that their post carries (with Ekman's emotions).
Our tool can then be used to pre-annotate each post by proposing automatically an emotion. After posts have been
annotated and published to the forum, it is possible to filter the existing posts by their annotated emotion and as
such find all the posts that carry sadness for example.

\subsubsection{Global opinion of TV viewers}

An early application for sentiment analysis has been the annotation of movie reviews in order to automatically infer
the sentiment of viewers towards a movie \cite{pang2002}. We propose to apply the same idea to the TV shows. It is
known that regular TV shows have Twitter fan-base who discusses the show. The idea is thus to conduct
sentiment/emotion analysis on Twitter streams that are related to a particular TV show. The ongoing work related to
that application is thus inline with recent work in sentiment analysis and emotion detection in Twitter, see for
instance \cite{mohammad2012}.

\section{Conclusion}

When developing a sentiment/emotion analysis service that is meant to be generic enough to work with several
different applications, it is important to consider whether the algorithms are tied to a particular domain, whether
the representation of output emotions is homogenous for all applications and whether the algorithms may be adapted
to other languages than English. We detailed these three problems while mentioning the existing solutions to them.
We introduced our own sentiment/emotion analysis service developed in the context of the Empathic Products ITEA
project which partially adresses these problems. While interoperability seems satisfactory enough and multilinguality
support has already been shown to be robust when using machine translation, the domain dependence aspects could be
improved. In particular we evaluated the algorithms on two quite different domains, the news headlines provided by
the Semeval-07 affective task evaluation and the tweets provided by the Semeval-13 sentiment analysis in Twitter
evaluation. The results show significant difference, probably caused by the difference of text length between the two
types of dataset. Nevertheless, for future work we are considering approaches that are more hybrid such as
\cite{andreevskaia2008} in order to tackle domain dependence.\\

\emph{This work was conducted in the context of the ITEA2 "Empathic Products" project, ITEA2 1105, and is supported by
funding from the French Services, Industry and Competitivity General Direction.}

%\bibliographystyle{splncs}
%\bibliography{bibli}

\end{document}